# VTQA: Visual Text Question Answering via Entity Alignment and Cross-Media Reasoning

Kang Chen
Harbin Institute of Technology
chenkangcs@stu.hit.edu.cn

Xiangqian Wu
Harbin Institute of Technology
xqwu@hit.edu.cn

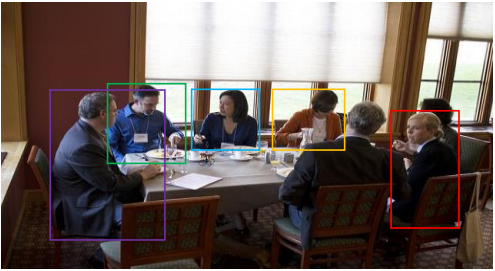

**Figure 1: Example in our dataset with the question-answer pairs and their corresponding image and text. Different representations of the same object in text and image are identified with the same color. For example, 'Elena' in the text and the object bounding box corresponding to 'Elena' in the image are marked red.**

## ABSTRACT

The ideal form of Visual Question Answering requires understanding, grounding and reasoning in the joint space of vision and language and serves as a proxy for the AI task of scene understanding. However, most existing VQA benchmarks are limited to just picking the answer from a pre-defined set of options and lack attention to text. We present a new challenge with a dataset that contains 23,781 questions based on 10124 image-text pairs. Specifically, the task requires the model to align multimedia representations of the same entity to implement multi-hop reasoning between image and text and finally use natural language to answer the question. The aim of this challenge is to develop and benchmark models that are capable of multimedia entity alignment, multi-step reasoning and open-ended answer generation.

## 1 INTRODUCTION

A long-term goal of AI system is to understand the complex real world like human beings, and Question Answering (QA) is a good task to evaluate the understanding ability of the AI system. To answer questions, people need to extract information from multiple modalities, such as text, images and structured data like knowledge base, graphs and tables. And furthermore, people need to align the information and do multi-steps reasoning between different modalities.

Although Visual Question Answering (VQA) [1] has been widely researched as a multimedia QA task, VQA models only extract information from image when answering questions and focus mainly on scene recognition, counting, color and other visual detection tasks, which do not require much logical reasoning or assignment between different modalities. Only recently, there are some attempts to introduce more modal information into VQA tasks. For example: (1) FVQA [2] and OK-VQA [3] combine knowledge base (KB) with VQA task, which requires the ability of knowledge understanding and multi-step reasoning. But it is difficult to construct a comprehensive KB in real world, which makes it restricted to answer the open-ended questions; (2) TextbookQA [4] and ScienceQA [5] use textbook as data sources, involving texts, images, tables and other multimodal information. But most images in these datasets are manually drawn schematic diagrams, and the questions are all in a multiple-choice setting, which are far from the real world.

To address these issues, more recently, there appears some new datasets such as MultiModalQA [6] and MuMuQA [7], which involve reasoning across texts, images and so on. However, these datasets are all conducted with extracted QA. And for MultiModalQA dataset, each image is associated with a Wikipedia entity, therefore the reasoning of images essentially degenerates to ranking the images and using the entities correspond to the top-ranked images, which weakens the need for cross-media reasoning ability. As for MuMuQA, although it requires grounding between image and text along with multi-hop reasoning, there are still some problems: (1) data are all from news, resulting in most questions related to human beings; (2) questions all follow a specific pattern: first perform image entity grounding and then find the answer in the news body text; (3) there is only 1384 human-curated examples and the training data is automatically generated, which leads to poor quality and difficult to train. Current multimedia benchmarks are still far from the real-world QA scene and cannot measure the multimedia understanding ability of AI system well.

To bridge this gap, we introduce a new challenge named "Visual Text Question Answering" (VTQA) along with a corresponding dataset, which includes only questions that require multi-hop reasoning through both image and text. All the annotations in VTQA dataset are first marked in Chinese, and we will also provide the corresponding English translations. In this challenge, we will divide the English and Chinese tracks and rank them respectively.

To answer VTQA questions, the proposed model needs to: (1) learn to identifying entities in image and text referred to the question, (2) align multimedia representations of the same entity, and (3) conduct multi-steps reasoning between text and image and output open-ended answer. The VTQA dataset consists of 10124 image-text pairs and 23,781 questions. The images are real images from MSCOCO dataset [8], containing a variety of entities. The annotators are required to first annotate relevant text according to the image, and then ask questions based on the image-text pair, and finally answer the question open-ended. We promise the richness of text information by limiting the minimum length of text (more than 100 Chinese characters) and remove the questions that can be answered only by image or text to ensure the complexity of questions.

In this challenge, the model is expected to answer the question according to the given image-text pair. Information diversity, multimedia multi-step reasoning and open-ended answer make our task more challenging than the existing dataset. The aim of this challenge is to develop and benchmark models that are capable of multimedia entity alignment, multi-step reasoning and open-ended answer generation.

## 2 CHALLENGE TASK

As illustrated in Figure 1, given an image-text pair and a question, a system is required to answer the question by natural language. Importantly, the system needs to: (1) analyze the question and find out the key entities, (2) align the key entities between image and text, and (3) generate the answer according to the question and aligned entities. For example, in Figure 1, the key entity of Q1 is "Elena". According to the text "gold hair", we can determine that the second person from the right in the image is "Elena". Finally, we further answer "suit" based on the image information. As for Q2, which is a more complex question, the previous steps need to be repeated several times to answer it.

## 3 DATASET

### 3.1 Dataset Construction

To promote the progress of this open-ended multimedia multi-hop VTQA challenge, we collect a new dataset. Our dataset consists of 10124 image-text pairs and 23,781 questions. We collect data through newly developed annotation interface. Images from MSCOCO dataset are shown in the interface and annotators are first asked to write a paragraph of text related to the image. The annotators then come up with questions based on the image-text pair and the annotation process requires that the questions cannot be answered only by image or text. Finally, the other annotators mark the answers to the questions and indicate the categories of the answers. We set three categories for answers: (1) 'YN' means yes-or-no answer, (2) 'E' means that the answer is extracted from the text, and (3) 'G' means that the answer is generated from the text-image pair. And the annotators need to label 'yes' or 'no' in English for the yes-or-no answer, since there are too many words to express 'yes' or 'no' in Chinese, for example, '可以' and '是的' both mean 'yes'.

### 3.2 Dataset Splits

We randomly split the dataset into training, validation, test-dev and test splits and each image will only appear in one split. Each split has 11312, 1245, 2189, 9035 samples, respectively. The training, validation, test-dev and test splits are provided as JavaScript Object Notation (JSON) formatted text files called train.json, val.json and test.json, where each data sample is stored as a dictionary. The attributes in the JSON file are as follows.

- qid: Unique id for each question.
- image_local_path: Source path for the image in the dataset
- text: Annotated text based on the image.
- question: Annotated question according to the image-text pair.
- answer: Annotated open-ended answer, usually a short phrase.
- answer_type: Answer type, which is described in Section 3.1.
- yes_or_no: Only work for the yes-or-no answers, where 'yes' for the positive answer and 'no' for negative answer. Will not appear in other types of answers.

### 3.3 Dataset Analysis

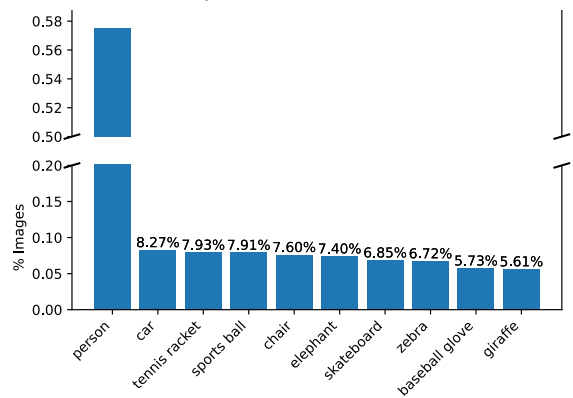

Figure 2: Top-10 categories distribution of the images used in our dataset.

Images in this dataset are from MSCOCO dataset, which contains multiple objects and rich contextual information. Figure 2

presents the top-10 categories distribution of the images used in this dataset. Unlike MuMuQA, in which most images are related to people, images in our dataset contain more kinds of objects.

The texts and questions length statistics are shown in Figure 3. We see that most texts range from 300 to 600 and most questions range from 9 to 21. Obviously, compared with the previous VQA datasets (usually less than 20 words), our dataset also puts forward higher requirements for text understanding to extract information from the long text.

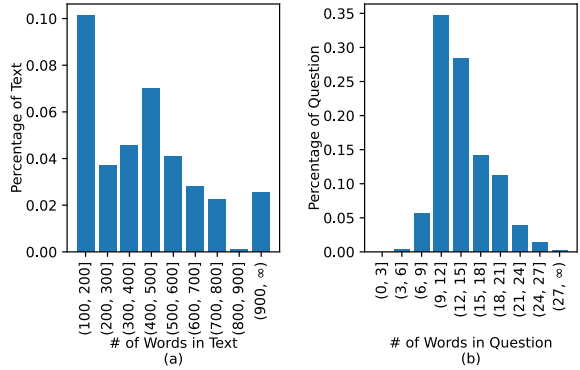

**Figure 3: Percentage of questions and texts with different word lengths.**

## 4 EVALUATION CRITERIA

As the answers divided into three types, we use different metrics for distinct types of answers.

**Exact match (EM)**. This metric measures the percentage of predictions that match the ground truth answer exactly and will be used in all types of answers.

**(Macro-averaged) F1 score (F1)**. This metric measures the average overlap between the prediction and ground truth answer. We treat the prediction and ground truth as bags of tokens and compute their F1. This metric will be used for the 'E' and 'G' types of answers.

**YN accuracy (YNAcc)**. This metric is only used for the 'YN' type of answers. The answer will be transformed into 'yes' or 'no' by a pre-defined yes-or-no dictionary. Then we calculate the accuracy just based on the yes-or-no answer.

The EM metric is used for ranking models and the other metrics are presented to show the performance of the algorithm and are not used for ranking.

All training and verification data will be published online for participants to train their models. There are two data split for participants to eval their models: (1) Test-dev set, which will be published without corresponding answers. Participants can predict answers of the test-dev set, generate, and submit the JSON files to quickly evaluate their own models. (2) Test set, which will be private, and the final ranking of this challenge will be based on the overall EM score of the test set. To evaluate models in test set, participants need to use a Docker image as submission. We will provide baseline implementations to minimize the overhead for participants. We will evaluate the models in test split and rank the participant models by the overall EM metric.

## 5 Baseline

In this section, we describe a competitive baseline method for evaluation on our benchmark, which is called Key Entity Cross-Media Reasoning Network (KECMRN). Before presenting the KECMRN, we first introduce its basic component, the KECMR layer. The KECMR layer is a modular composition which consists of one Key Entity Extract (KEE) layer and multiple Cross-Media Reason (CMR) layers. The KEE layer and CMR layer are composed of attention unit and feed-forward unit from [9]. Then we use the KECMR module with other layers to combine our KECMRN. Finally, we evaluate the KECMRN on our VTQA dataset.

### 5.1 Attention and Feed-Forward Units

As shown in [9], the combination of attention unit and feed-forward unit has strong representational and learning ability. We use the same settings to construct our units.

**Multi-Head Scaled Dot-Product Attention Unit.** The input of scaled dot-product attention consists of queries and keys of dimension $d_k$, and values of dimension $d_v$. We calculate the dot products of the query with all keys, divide each by $\sqrt{d_k}$ and apply a softmax function to obtain the attention weights on the values. Given a query $q \in \mathbb{R}^{1\times d}$, $n$ key-value pairs (packed into a key matrix $K \in \mathbb{R}^{n\times d}$ and a value matrix $V \in \mathbb{R}^{n\times d}$), the attended feature is obtained as:

$$Attention(Q,K,V) = softmax(\frac{QK^T}{\sqrt{d_k}})V \quad (1)$$

Multi-head attention divides the input into $h$ parts and makes single attention on each part. The attended feature is given by:

$$MH(Q,K,V) = Concat(head_1, ..., head_h)W^O \quad (2)$$
$$head_i = Attention(QW_i^Q, KW_i^K, VW_i^V) \quad (3)$$

where $W_i^Q, W_i^Q, W_i^Q \in \mathbb{R}^{d\times d_h}$ and $W^O \in \mathbb{R}^{h*d_h\times d}$ are the projection matrices.

**Feed-Forward Unit.** The feedforward unit takes the output features of the multi-head attention layer, and further transforms them through two fully-connected layers with ReLU activation in between.

$$FFN(x) = \max(0, xW_1 + b_1)W_2 + b_2 \quad (4)$$

While the linear transformations are the same across different positions, they use different parameters from layer to layer.

### 5.2 Key Entity Extract Layer

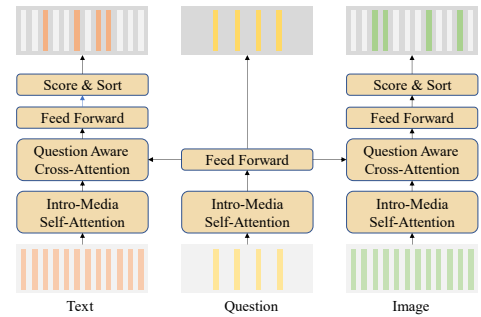

**Figure 4: Key Entity Extract Layer.**

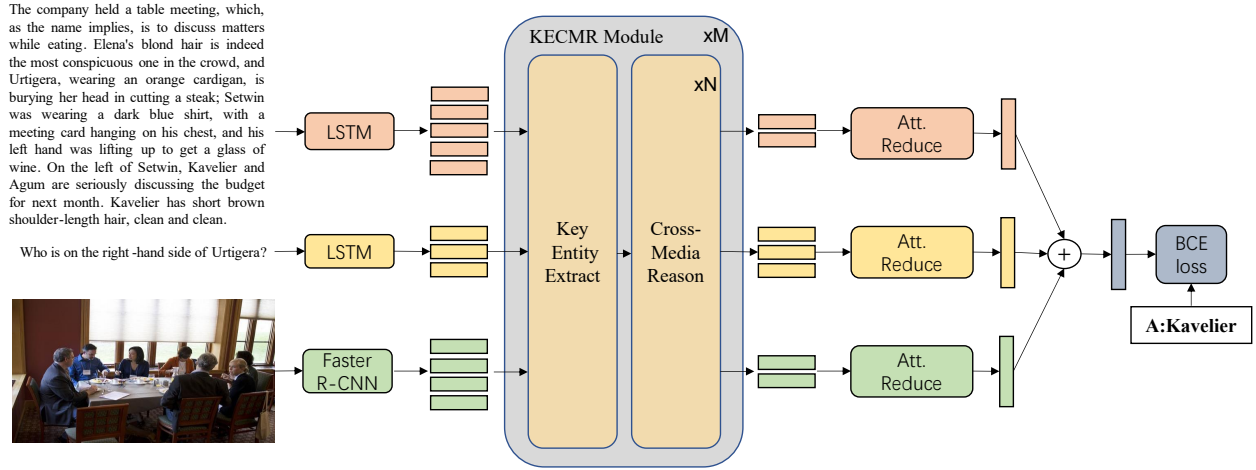

Figure 5: Overall flowchart of the Key Entity Cross-Media Reasoning Network.

As stated in Section 2, the first step to answer VTQA questions is to find out the key entity according to the questions. We compose the attention unit and feed-forward unit to integrate question information into the text and the image respectively. Then we apply a fully-connected layer to the question-aware text/image features to get the score for each feature. Finally, we extract the top-k features as key entities.

The complete KKE layer is shown in the Figure 4. Given input text features $T \in \mathbb{R}^{l_t \times d}$, image features $I \in \mathbb{R}^{l_i \times d}$ and question features $Q \in \mathbb{R}^{l_q \times d}$, the KEE layer can be formulated by:

$$Q = FFN(MH(Q,Q,Q))$$
$$T = FFN(MH(MH(T,T,T),Q,Q))$$
$$I = FFN(MH(MH(I,I,I),Q,Q)) \quad (4)$$
$$score_T = W_T T + b_T$$
$$score_I = W_I I + b_I$$

The final score will be used to sort the image/text features and the top-k features are considered as image/text key entities. To unify the expression, we treat all the question features as key entities.

### 5.3 Cross-Media Reason Layer

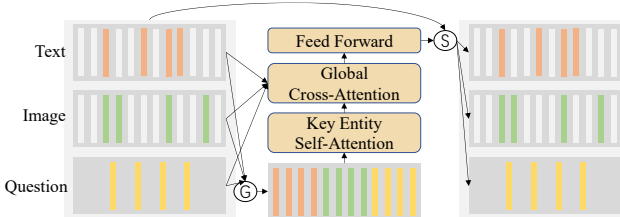

Figure 6: Cross-Media Reason Layer. 'G' means gather and 'S' means scatter.

The CMR layer is designed for multi-step reasoning across medias. As show in Figure 6, we first gather the key entities as $S \in \mathbb{R}^{l_k \times d}$ from input features. Then the key entities pass through self-attention, global cross-attention based on original features and feed-forward unit. Finally, the key entities are scattered to the input features. As the requirement for multi-steps reasoning, we inserted multiple CMR layers in a KECMR module. The CMR layer can be formulated by:

$$S = gather([T,I,Q])$$
$$S = FFN(MH(MH(S,S,S),[T,I,Q],[T,I,Q])) \quad (4)$$
$$[T,I,Q] = scatter(S,[T,I,Q])$$

### 5.4 Key Entity Cross-Media Reason Network

We use the same network framework as [10] but replace MCA layer with our KECMR module and add an extra text stream. As show in Figure 5, The input image is represented as a set of regional visual features in a bottom-up manner [11]. The input question and text are transformed to features by passing through a one-layer LSTM network [12]. Then we use our KECMR module several times to extract key entities and conduct multi-steps cross-media reasoning. Finally, we use the attention reduce layer to fuse the multimedia features and project the fused feature into the answer probability distribution.

### 5.5 Baseline Results

The hyper-parameters of our model used in the experiments are as follows. The dimensionality of input image features $d_i$, input question features $d_q$, input text features $d_t$ and fused multimodal features $d_z$ are 2,048, 512, 512, and 1,024, respectively. The latent dimensionality d in the multi-head attention is 512, the number of heads h is set to 8, and the latent dimensionality for each head is dh = d/h = 64. The number of CMR layer in each KECMR module and the number of KECMR module are 2 and 2.

We train our model on the *train* set for 13 epochs and evaluate our model both in *test-dev* and *test* sets. The results are shown in table 1.

Table 1: Results of our KECMRN in test-dev and test sets.

| Dataset | EM | YN-Acc | E-F1 | G-F1 |
|---|---|---|---|---|
| Test-dev | 0.527 | 0.782 | 0.619 | 0.527 |
| Test | 0.513 | 0.776 | 0.605 | 0.511 |

## 6 ADMINISTRATIVE DETAILS

A challenge website[1] has been setup with a commitment to be maintained at least for the next 3 years and the dataset will be continuously updated to keep our challenge competitive in the field of multimedia-QA. We will do our best to work with ACM Multimedia Conference organizers to publicize the Grand Challenge tasks to researchers for participation. The organizers and the Programme Committee members are listed below, and all the members listed in Programme Committee have already agreed to serve as reviewers for papers submitted to the Challenges track. Organizers.

- Wu, Xiangqian. xqwu@hit.edu.cn. Harbin Institute of Technology.
- Chen, Kang. chenkangcs@stu.hit.edu.cn. Harbin Institute of Technology.
- Zhao, Tianli. tlzhaoipir@stu.hit.edu.cn. Harbin Institute of Technology.

Programme Committee.
- Che, Wanxiang. Harbin Institute of Technology.
- Chen, Kang. Harbin Institute of Technology.
- Cui, Yiming. Joint Laboratory of HIT and iFLYTEK Research.
- Deng, Ke. Tsinghua University.
- Ding, Xiao. Harbin Institute of Technology.
- Fan, Xiaopeng. Harbin Institute of Technology.
- Feng, Xiaocheng. Harbin Institute of Technology.
- Fu, Ruiji. MMU KuaiShou Inc.
- Gao, Yang. Beijing Institute of Technology.
- Hong, Xiaopeng. Harbin Institute of Technology.
- Li, Juanzi. Tsinghua University.
- Li, Piji. Nanjing University of Aeronautics and Astronautics.
- Li, Yang. Northeast Forestry University.
- Nie, Liqiang. Harbin Institute of Technology.
- Su, Jinsong. Xiamen University.
- Sun, Maosong. Tsinghua University.
- Tang, Duyu. Tencent AI Lab.
- Wu, Xiangqian. Harbin Institute of Technology.
- Xia, Rui. Nanjing University of Science and Technology.
- Xu, Tong. University of Science and Technology of China.
- Yang, Liang. Dalian University of Technology.
- Yang, Min. Shenzhen Institute of Advanced Technology.
- Zan, Hongying. Zhengzhou University.
- Zhao, Tianli. Harbin Institute of Technology.

---

[1] https://visual-text-QA.github.io/